\documentclass[10pt,journal,letterpaper,twoside,compsoc]{IEEEtran}

\usepackage{times}

\usepackage{epsfig}
\usepackage{graphicx}
\usepackage[belowskip=-10pt,aboveskip=2pt,font=small]{caption}
\usepackage{dblfloatfix}

\usepackage{amsmath}
\usepackage{amssymb}

\usepackage{multirow}
\usepackage{rotating}
\usepackage{booktabs}
\usepackage{algpseudocode}
\usepackage{algorithm}

\usepackage{enumerate}
\usepackage[olditem,oldenum]{paralist}

\usepackage{color}
\usepackage{comment}
\definecolor{dblue}{rgb}{0,0,0.5}
\usepackage[pagebackref=true,breaklinks=true,letterpaper=true,colorlinks,bookmarks=false,citecolor=dblue,linkcolor=blue]{hyperref}

\definecolor{orange}{RGB}{225, 90, 0}
\definecolor{teal}{RGB}{5, 210, 150}
\definecolor{yellow}{RGB}{220, 210, 10}
\definecolor{purple}{RGB}{100, 0, 205}

\DeclareMathOperator*{\argmax}{argmax}

\begin{document}


\title{Exploring Nearest Neighbor Approaches for Image Captioning}

\author{Jacob Devlin, Saurabh Gupta, Ross Girshick, Margaret Mitchell, C. Lawrence Zitnick
\IEEEcompsocitemizethanks{
\IEEEcompsocthanksitem  J. Devlin, R. Girshick, M. Mitchell , and C. L. Zitnick are with Microsoft Research, Redmond.
\IEEEcompsocthanksitem Saurabh Gupta is with University of California, Berkeley.}
}



\IEEEcompsoctitleabstractindextext{\begin{abstract}
We explore a variety of nearest neighbor baseline approaches for image captioning. These approaches find a set of nearest neighbor images in the training set from which a caption may be borrowed for the query image. We select a caption for the query image by finding the caption that best represents the ``consensus'' of the set of candidate captions gathered from the nearest neighbor images. When measured by automatic evaluation metrics on the MS COCO caption evaluation server, these approaches perform as well as many recent approaches that generate novel captions. However, human studies show that a method that generates novel captions is still preferred over the nearest neighbor approach.
\end{abstract}}

\maketitle
\section{Introduction}

The automatic generation of captions for images has recently received significant attention
\cite{ryan2014multimodal,mao2014explain,vinyals2014show,karpathy2014deep2,kiros2014unifying,donahue2014long,fang2014captions,chen2014learning,lebret2015phrase,lebret2014simple,lazaridou2015combining,xuAttend2015}. This surge in research is due in part to the creation of large caption datasets \cite{grubinger2006iapr,ordonez2011im2text,hodosh2013framing,young2014image,jianfu2015,COCO}, and new learning techniques \cite{Hinton,hochreiter1997long}.  Recently proposed methods for caption generation share many similarities, including the use of deep learned image features \cite{Hinton,jia2014caffe,simonyan2014very}, and language models using maximum entropy \cite{fang2014captions}, recurrent neural networks \cite{chen2014learning,karpathy2014deep2}, and LSTMs \cite{vinyals2014show,donahue2014long}. An integral feature of all these methods is their ability to generate novel captions.

We seek to better understand how important the generation of {\it novel} captions is for the task of automatic image captioning when using the benchmark MS COCO dataset \cite{COCO}.  Previously, several papers proposed producing image captions by first finding similar images, and then copying their captions \cite{farhadi2010every,ordonez2011im2text,hodosh2013framing}. Given larger caption datasets such as the MS COCO \cite{COCO} dataset, which contains 100,000s of captions, the chances of finding an appropriate caption may increase, making such approaches more useful.  Vinyals et al.~\cite{vinyals2014show} found that up to $80\%$ of the captions generated by their approach were identical to captions in the MS COCO training dataset, while still achieving near state-of-the-art results. This provides evidence that copying captions may indeed achieve good results.  However, if the images in the MS COCO dataset contain too much diversity, or capture many rare occurrences, approaches that copy captions directly may not perform as well as those those that can additionally generate novel captions.

\begin{figure}
  \centering
  \includegraphics[width=\linewidth]{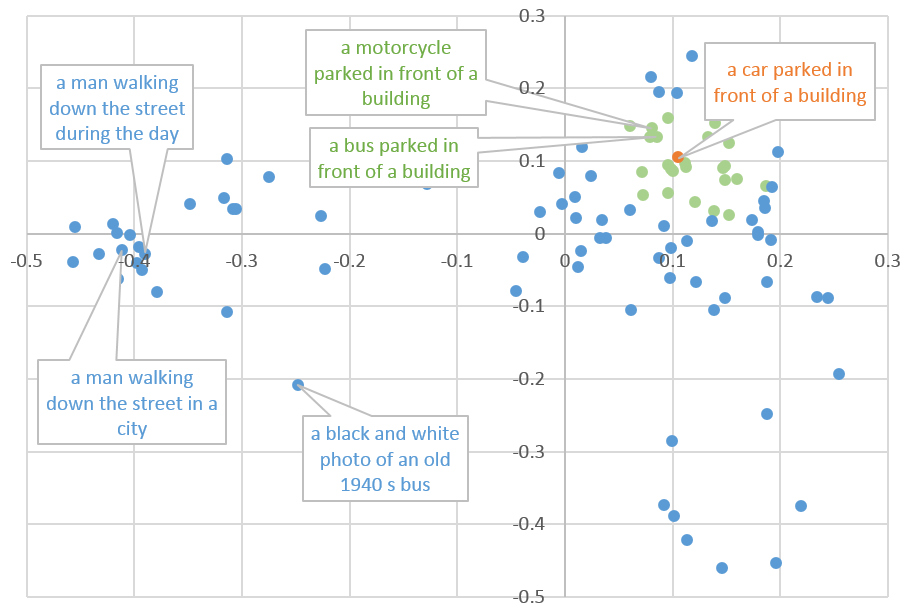}\\
  \caption{Example of the set of candidate captions for an image, the highest scoring $m$ captions (green) and the consensus caption (orange). This is a real example visualized in two dimensions.}\label{fig:consensus}
\end{figure}


In this paper we expand on \cite{devlin2015language} by providing a detailed exploration into nearest neighbor (NN) approaches for image captioning. Nearest neighbor approaches have a rich history in work on predicting words given images, used in face recognition \cite{PhillipsEtAl02} as well as recent work in retrieval-based caption generation \cite{farhadi2010every,mason2014}.  We focus on nearest neighbor approaches to gain further insight into the limitations of the captioning task, and to explore the properties of the largest captioning dataset to date, the MS COCO dataset. We hope to provide context for the recent advances in this area \cite{ryan2014multimodal,mao2014explain,vinyals2014show,karpathy2014deep2,kiros2014unifying,donahue2014long,fang2014captions,chen2014learning,lebret2015phrase,lebret2014simple,lazaridou2015combining,xuAttend2015}.

Our nearest neighbor approach finds a set of $k$ nearest images.  Which images are ``nearest'' can be defined in several ways, and we examine using GIST \cite{GIST}, pre-trained deep features \cite{simonyan2014very}, and deep features fine-tuned for the task of caption generation \cite{fang2014captions}. Once a set of $k$ NN images are found, the captions describing these images are combined into a set of candidate captions from which the final caption is selected, Figure \ref{fig:consensus}. We select the best candidate caption by finding the one that scores highest with respect to the other candidate captions.  We refer to this as the ``consensus'' caption.  The scores between pairs of captions are computed using either the CIDEr \cite{cider} or BLEU \cite{bleu} metric.

Surprisingly, we find that this simple NN approach outperforms many novel caption generation approaches as measured by BLEU \cite{bleu}, METEOR \cite{meteor} and CIDEr \cite{cider} on the MS COCO testing \cite{COCO} dataset. We find that using simple features for finding nearest neighbors such as GIST \cite{GIST} do not perform well. However, deep features, especially those fine-tuned specifically for caption generation \cite{fang2014captions} are very effective at finding  images from which high-scoring captions may be borrowed. While the NN approaches perform well when evaluated using automatic metrics, a crowdsourced study shows that humans still prefer a system that generates novel captions \cite{fang2014captions} by a significant margin. Further human studies still need to be performed to see how the NN approaches compare to other generation-based approaches.

\begin{figure*}
  \centering
  \includegraphics[width=\linewidth]{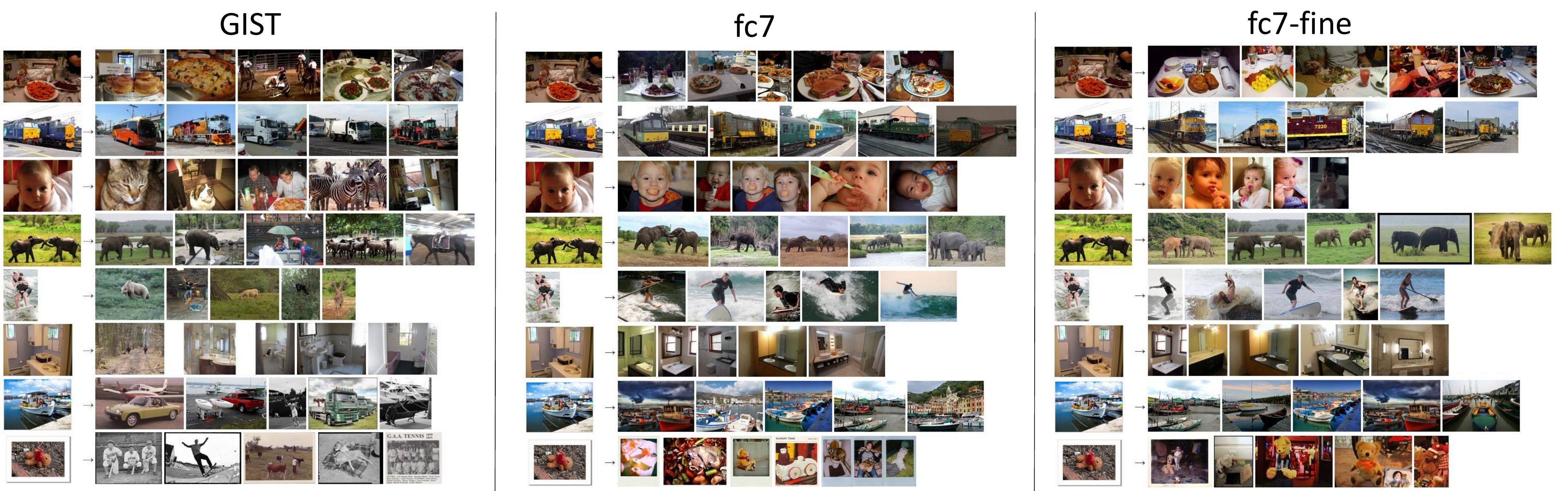}\\
  \caption{Example nearest neighbor matches using different feature spaces.}\label{fig:NN}
\end{figure*}

\section{Related Work}

Several early papers proposed producing image captions by copying captions from other images \cite{farhadi2010every,ordonez2011im2text,hodosh2013framing,Kannan:2014}. \cite{farhadi2010every} use nearest neighbors to define image and caption features, capturing information about objects, actions, and scenes, where \cite{ordonez2011im2text} use a combination of object, stuff, people and scene information.  \cite{mason2014} use GIST nearest neighbors to the query image.  \cite{hodosh2013framing} use Kernel Canonical Correlation Analysis to map images and captions to a common space where the nearest caption can be found.
While not using explicit captions, \cite{Kannan:2014} explores the task of captioning images using surrounding text on webpages.


Hodosh et al.~\cite{hodosh2013framing} popularized the task of image and caption ranking. That is, given an image, rank a set of captions based on which are most relevant. They argued that this task was more correlated with human judgment than the task of novel caption generation measured using automatic metrics. Numerous papers have explored the task of caption ranking \cite{mao2014explain,socher2013grounded,frome2013devise,karpathy2014deep,chen2014learning,maoDeepCaptioning}. These approaches could also be used to rank the set of training captions, and used to select the one that is most relevant. As far as we are aware, how well such an approach would perform on the MS COCO caption dataset for generation is still an open question. In this paper, we only explore a simple nearest neighbor baseline approach.

\section{Approach}

In this section, we describe our set of approaches for image captioning. We assume a dataset of training images with a set of corresponding captions. We use the MS COCO \cite{COCO} training dataset containing $82,783$ images with 5 captions each, for a total of 413,915 captions. In our approach \cite{devlin2015language} we first find a set of $k$ NN images in the training dataset. The consensus caption returned by our approach is selected from the set of candidate captions describing the set of $k$ NN training images.

\subsection{Nearest Neighbor Images}

Our first task is to find a set of $k$ nearest training images for each query image based on visual similarity.\footnote{The value of $k$ is chosen optimally for each feature set, and typically ranges from 50-200.} We find the $k$ NNs using cosine similarity with the following feature spaces:
\begin{itemize}
\item {\bf GIST}: We use the popular approach of \cite{GIST} to compute a set of global image features based on the summation of low-level image features, such as contours or textures. GIST is computed on images resized to $32 \times 32$ pixels.
\item {\bf fc7}: Our first set of deep features are computed using the {\tt fc7} layer of the VGG16 Net \cite{simonyan2014very}. The network was trained using the 1,000 ImageNet classification task \cite{Imagenet}. The features are computed using a single window with resolution $224\times224$. Images are rescaled to make the longer side 224 pixels. Empty image regions were replaced by the mean image.
\item {\bf fc7-fine}: These features are computed in the same manner as {\tt fc7}. However, the weights of the VGG16 network are fine-tuned for the image captioning task. Specifically, its weights are initialized using the ImageNet task, and the weights are fine-tuned on the task of classifying the 1,000 most commonly occurring words in image captions \cite{fang2014captions}.
 \end{itemize}

The image features are computed for every image in the training dataset. The neighbor images are found by exhaustively computing the cosine similarly between the query image and the training images. In Figure \ref{fig:NN}, we show several examples of NN matches using different feature spaces. Notice how the NNs found using deep features are more semantically similar.

\subsection{Consensus Caption}

Given $k$ nearest training images for a given test image, we take the union of their captions to create a set $C$ of $n$ candidate captions. Our task is to select the ``best'' or consensus caption from the set $C$, as seen in Figure \ref{fig:consensus}. There are five captions per image in the MS COCO dataset, so $n = 5k$. 
We define the consensus caption $c^*$ as the one that has the highest average lexical similarity to the the other captions in $C$. This scoring function is:
\begin{equation}
c^* = \argmax_{c\in C} \sum_{c'\in C} \mathit{Sim}(c, c'),
\label{eqn:consensus}
\end{equation}
where $\mathit{Sim}(c, c')$ is the similarity score between two captions $c$ and $c'$. We explore two similarity functions: BLEU \cite{bleu}, which measures 1-to-4-gram overlap, and CIDEr \cite{cider}, which measures tf-idf weighted 1-to-4-gram overlap. Intuitively, this tf-idf weighting means that CIDEr pays more attention to rarer, more descriptive phrases. We use the CIDEr-D variant of CIDEr.

Some of the candidate captions in $C$ might be outliers and add noise to the computation of Equation (\ref{eqn:consensus}). A solution to this problem is to compute Equation (\ref{eqn:consensus}) only over a subset $M$ of $C$, where the number $m$ of captions in $M$ is less than $n$. This can be thought of as finding the centroid of a large cluster of captions, as demonstrated in Figure~\ref{fig:consensus}.  Using $M$, our final consensus caption is:
\begin{equation}
c^* = \argmax_{c\in C} \max_{M \subset C} \sum_{c'\in M} \mathit{Sim}(c, c').
\label{eqn:m}
\end{equation}
The inner maximization is over all size-$m$ subsets $M$ of $C$.

Intuitively, the consensus caption is a \textit{single} caption from the training data that can be used to describe \textit{many} images that are visually similar to the test image. Ideally, then, this caption is likely to also be an adequate description of the test image.

If the NN images are diverse, one would expect the chosen caption to be more generic, while if the NN images are quite similar, the selected caption may be specific. The reason is that the descriptiveness of captions is a basic risk vs. reward trade-off: {\tt a red car} is a better description than {\tt a car} if the car is red, but a significantly worse description when the car is actually blue. As a result, the caption's detail is directly dependent on the diversity of the dataset used to gather candidate captions.

\begin{figure*}
  \centering
  \includegraphics[width=\linewidth]{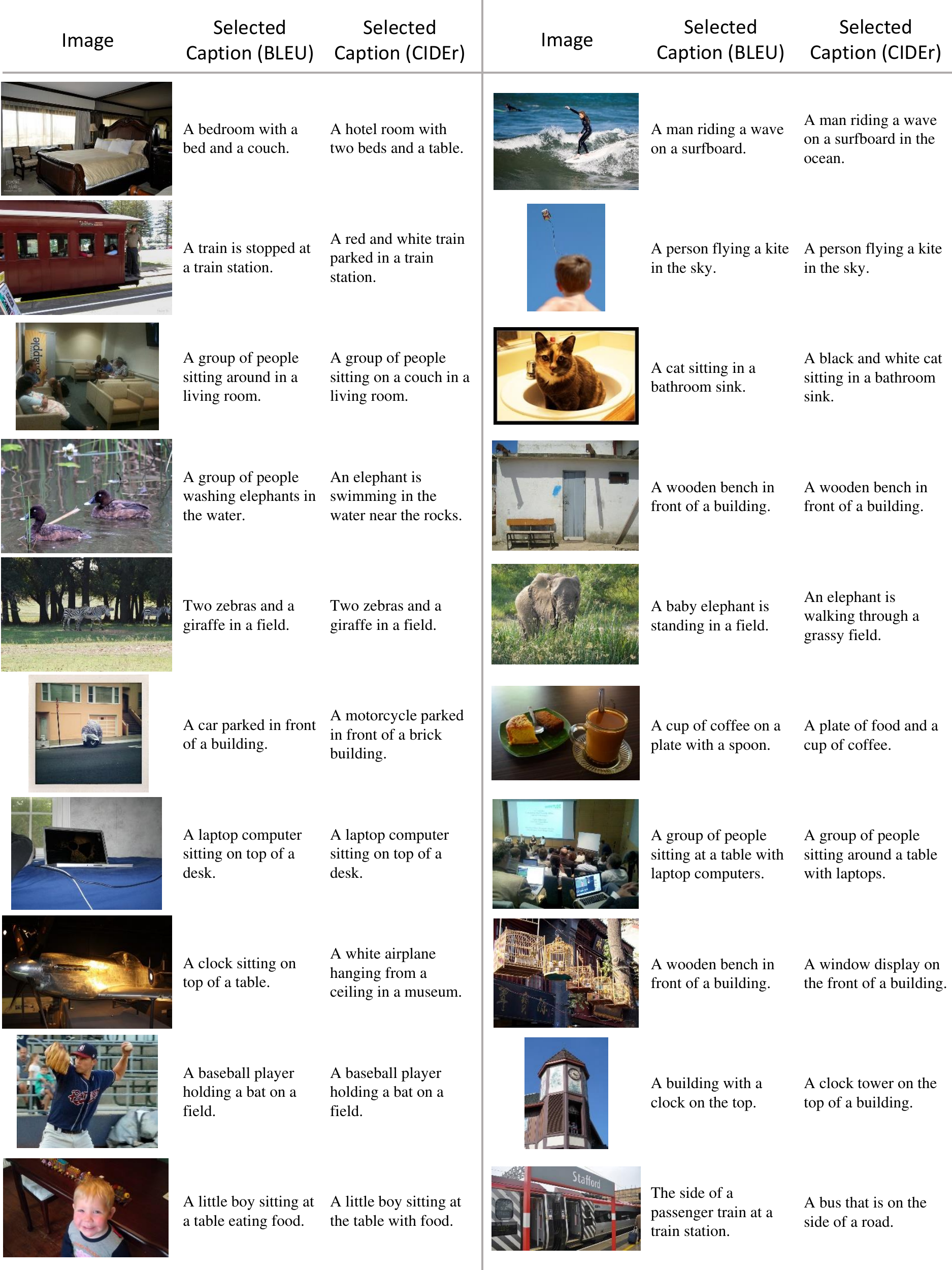}\\
  \caption{Several examples of randomly selected images and their selected consensus captions. The consensus caption is shown using the BLEU metric and CIDEr metric for scoring. Notice the chosen captions using CIDEr are more detailed.}\label{fig:qual}
\end{figure*}

In Figure, \ref{fig:qual} we show several examples of consensus captions using both CIDEr and BLEU. Subjectively, we can see that the CIDEr-tuned captions tend to be more descriptive than the BLEU-tuned captions, which is likely due to CIDEr's preference for rarer n-grams.

\section{Results}
\label{sec:results}

In this section, we provide results for several variants of the NN approach. For all our experiments, we use 82,783 MS COCO training images for training, and split 40,504 validation set into two halves: a ``tuning'' set for hyperparameter optimization, and a ``testval'' set for reporting results.

We begin by exploring the effect of $k$ and $m$ on the final accuracy. Next, we explore the different feature spaces that may be used to find NN images. Finally, we perform human studies to evaluate how well NN approaches perform as judged by humans, and report results on the MS COCO testing set.

\subsection{The Number of Nearest Neighbors}

\begin{figure}
  \centering
  \includegraphics[width=\linewidth]{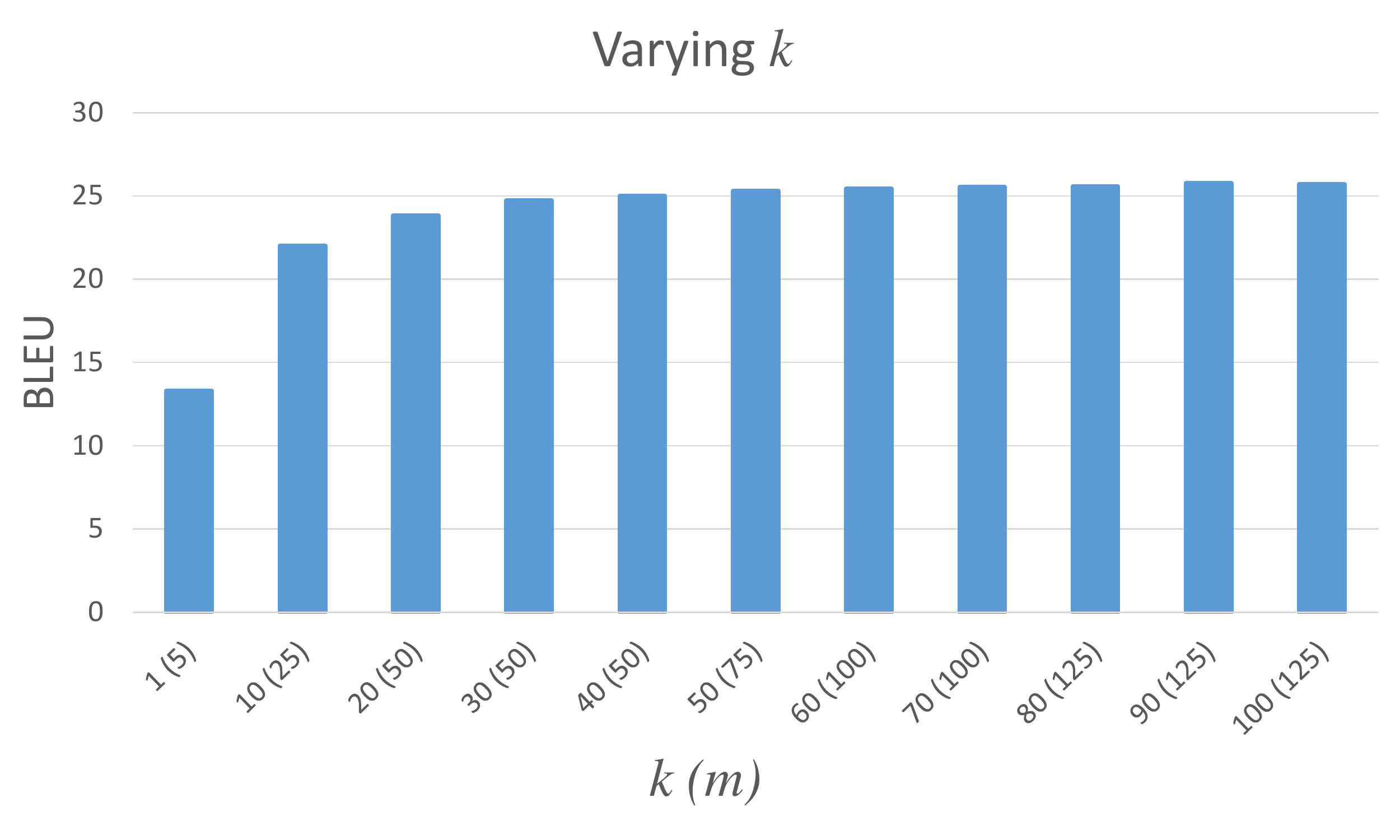}\\
  \caption{Resulting BLEU scores when varying number of NN images, $k$. The optimal $m$ for each $k$ is shown in parentheses.}\label{fig:K}
\end{figure}

How important is the selection of $k$, the number of nearest neighbor images used to create the candidate caption set $C$? In Figure \ref{fig:K}, we show BLEU scores \cite{bleu} as we vary $k$. Notice that for $k < 20$, significantly worse results are achieved. For reference, if only one image is selected and a caption is randomly chosen, the BLEU score is 11.2 \cite{donahue2014long}. For $k>60$ the BLEU scores are roughly similar.

\begin{figure}
  \centering
  \includegraphics[width=\linewidth]{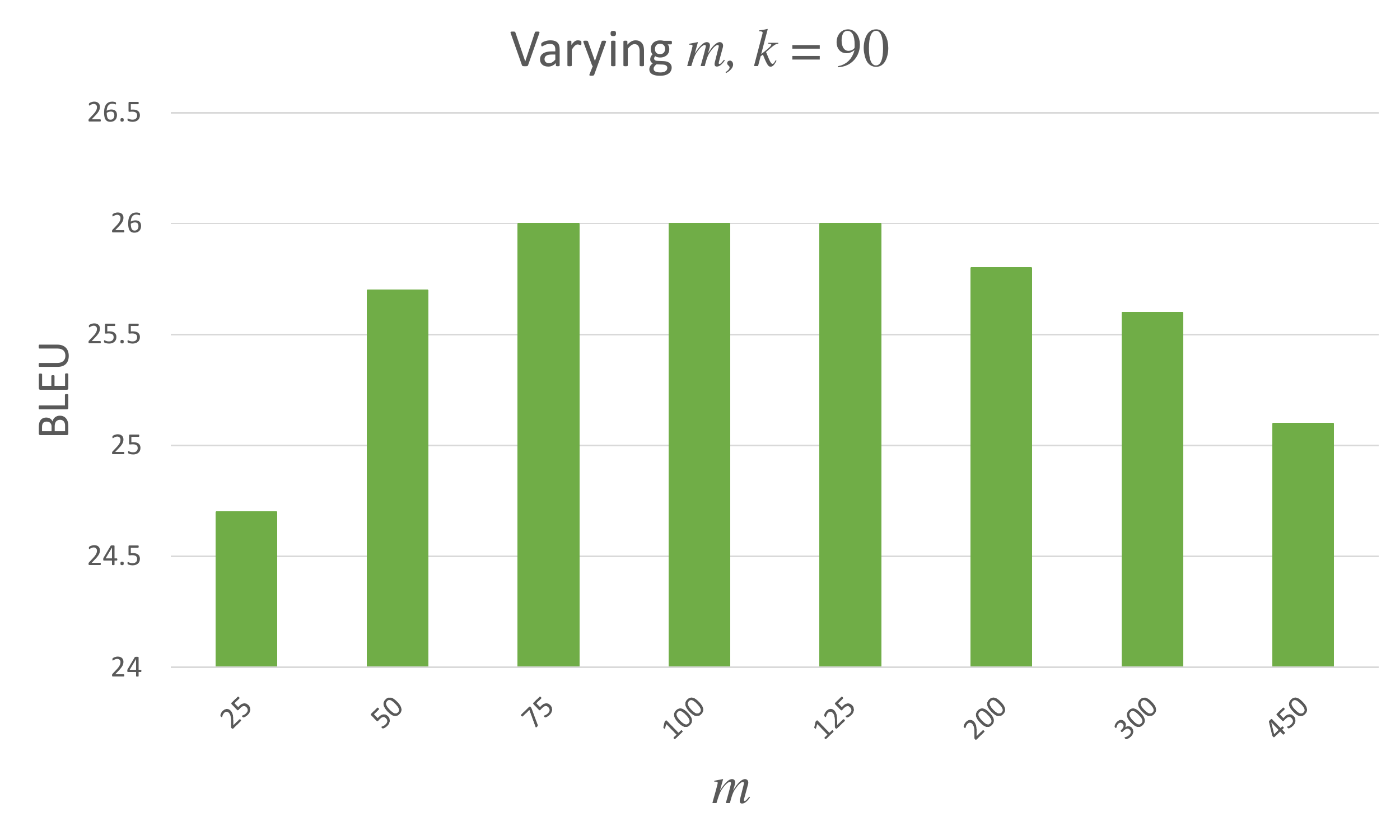}\\
  \caption{Resulting BLEU scores when varying the number of captions $m$ used to compute the consensus score. $k$ is held constant at 90.}\label{fig:M}
\end{figure}

In Figure \ref{fig:M}, we show results when we vary $m$, the number of candidate captions used to select the consensus caption (Equation \ref{eqn:m}). High scores are achieved for a variety of values for $m$ ranging between 50 and 200. If $m = n$, worse results are achieved, supporting our hypothesis that outlier captions should be removed.

As shown in Table \ref{tab:feature}, finding the best caption in Equation (\ref{eqn:m}) is slightly better using CIDEr ({\tt fc7-fine} (CIDEr))
 than BLEU ({\tt fc7-fine} (BLEU)). CIDEr performs better as measured by both CIDEr and METEOR. Not surprisingly, optimizing using BLEU preforms better when measured by BLEU.

\begin{table*}
\begin{center}
\begin{tabular}{l c c c  c c c}
\toprule
    & \multicolumn{3}{c}{c5} & \multicolumn{3}{c}{c40} \\

   Method & BLEU 4 & CIDEr & METEOR & BLEU 4 & CIDEr & METEOR \\
  \toprule

ME + DMSM \cite{fang2014captions} & \bf{29.1} &  \bf{0.912} & 24.7	&  \bf{56.7} &  \bf{0.925} & 33.1  \\
LRCN \cite{donahue2014long} & 27.7 &  0.869 & 24.2 & 53.4 & 0.891 & 32.2 \\
Vinyals et al. \cite{vinyals2014show} & 27.2 & 0.834 & 23.6 & 53.8 & 0.842 & 32.7 \\
Xu et al. \cite{xuAttend2015} & 26.8  & 0.850  & 24.3 & 52.3  & 0.878  & 32.3 \\
m-RNN \cite{maoDeepCaptioning} & 27.9 & 0.819 & 22.9 & 54.3 & 0.828 & 31.2 \\
MLBL \cite{ryan2014multimodal,kiros2014unifying} & 26.0 & 0.740 & 21.9 & 51.7 & 0.752 & 29.4 \\
NeuralTalk \cite{karpathy2014deep2} & 22.4 & 0.674 & 21.0 & 44.6 & 0.692 & 28.0 \\

\toprule
fc7-fine (CIDEr) & 27.9 (2) & 0.886 (2) & 23.7 (3) & 54.2 (2) & 0.916 (2) & 31.8 (5) \\

\toprule
Human & 21.7 & 0.854 &  \bf{25.2} & 47.1 & 0.910 &  \bf{33.5} \\

\bottomrule
\end{tabular}
\caption{Results on the MS COCO test set for c5 (left) and c40 (right). Best scores are shown in bold. Results on {\tt fc7-fine} (CIDEr) are shown, with its relative ranking compared to the automatic approaches shown in parentheses. For comparison, results using captions written by humans are also shown.}
\vspace{-5pt}
\label{tab:COCOtest}
\end{center}
\end{table*}

\subsection{Different Feature Spaces}

We now explore the effect of using different feature spaces. For these experiments, we use the values of $k$ and $m$ that produced the highest BLEU/CIDEr scores on the tuning half of the validation set. Results on the testval half of the validation set are shown in Table \ref{tab:feature}.\footnote{Results were computed on 4 references rather than 5 for consistency with \cite{fang2014captions}.}  GIST performs poorly since it doesn't capture the high-level semantics of the scenes as shown in Figure \ref{fig:NN}. The deeply learned features {\tt fc7} and {\tt fc7-fine} do significantly better. For comparison, we show the results of \cite{fang2014captions} (ME+DMSM) using a Maximum Entropy (ME) language model and a Deep Multimodal Similarity Model (DMSM). Interestingly, the {\tt fc7-fine} (BLEU) performs comparably to the ME+DMSM approach \cite{fang2014captions} as measured by BLEU.

How do these methods perform on query images that are {\it not} visually similar to the training data, versus images that are visually similar? To gain insight into this question, we compute the mean distance of the testval images to their 50 nearest training images based on {\tt fc7-fine} cosine distance. We then sort the mean distances and place the query testval images into ten bins, ranging from those with the closest NN images to those with the furthest.

\begin{figure}
  \centering
  \includegraphics[width=\linewidth]{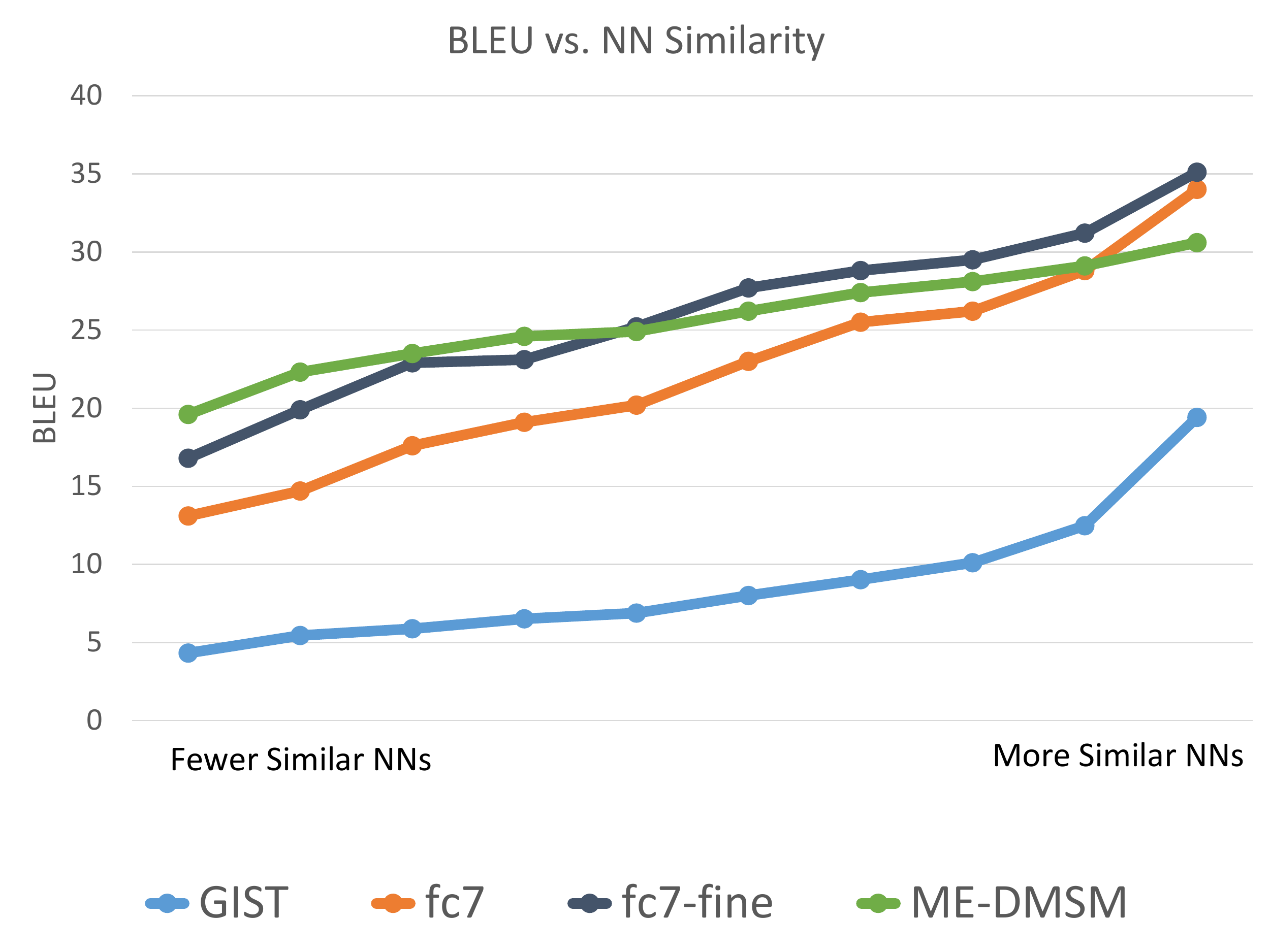}\\
  \caption{BLEU scores for various approaches when the testval images are split into 10 equally sized bins based on visual similarity to the training data. The bins are arranged from those with fewer close NNs (left), to those with more NNs images (right).}\label{fig:bins}
\end{figure}

The BLEU scores for each of these bins are shown in Figure \ref{fig:bins}. Unsurprisingly, the images that are most visually similar to the training achieve the highest BLEU scores across all approaches. However, compared to the generation-based approach of \cite{fang2014captions}, the nearest neighbor approaches perform better for highly similar images, but worse for highly dissimilar images. This suggests that the generation-based approach generalizes better to less common images, but doesn't do as well as borrowing captions for common images.

\begin{table}
\begin{center}
\begin{tabular}{l c c c c c}
\toprule
   Features & k & m & BLEU & CIDEr & METEOR \\
  \toprule

GIST & 80 & 100 & 9.0 & 0.23 & 12.2 \\
fc7 & 130 & 150 & 22.3	& 0.72 & 20.3 \\
fc7-fine (BLEU) & 90 & 125 & 26.0	&	0.85	&	22.5 \\
fc7-fine (CIDEr) & 80 & 200 & 25.1	&	0.90	&	22.8 \\
\toprule
ME + DMSM \cite{fang2014captions} & & & 25.7	&	0.92	&	23.6 \\
\bottomrule
\end{tabular}
\caption{BLEU \cite{bleu}, METEOR \cite{meteor} and CIDEr \cite{cider} scores on testval for NN approaches using different feature spaces. See text for descriptions of the feature spaces.}
\vspace{-5pt}
\label{tab:feature}
\end{center}
\end{table}

\begin{table}
\begin{center}
\begin{tabular}{l c c c c}
\toprule
    Approach & \multicolumn{3}{c}{Human Judgements} & BLEU \\
     \cline{2-4}\noalign{\smallskip}
    & Better & Equal & Better or Equal & \\
  \toprule

$k$-NN fc7-fine (BLEU) & 5.5\%	&	22.1\%	&	27.6\%  & 26.0\\
$k$-NN fc7-fine (CIDEr) &	6.3\%	&	20.2\%  & 26.5\% & 25.1 \\
ME + DMSM \cite{fang2014captions} & 7.8\% & 26.2\% & 34.0\% & 25.7 \\
\bottomrule
\end{tabular}
\caption{Results when comparing produced captions to those written by humans, as judged by humans. The percentage that are better than, equal to, and better than or equal to the captions written by humans are shown.}
\vspace{-5pt}
\label{tab:human}
\end{center}
\end{table}

\subsection{Human Evaluation}
An interesting question is how the captions selected by the NN approaches would perform when judged by humans. To explore this, we use the same experimental setup as \cite{fang2014captions}, in which human subjects are asked to judge whether a caption generated by a system is better than, worse than, or equal to a caption written by a human for that image. Each caption is evaluated 5 times and the majority is recorded. If a tie occurs (2-2-1), each of the top choices are given half a vote. The results are shown in Table \ref{tab:human}. The generation-based approach of \cite{fang2014captions} significantly outperforms the nearest neighbor approaches, despite the similar BLEU scores. As a point of reference, a baseline system from \cite{fang2014captions} that uses non-fine-tuned {\tt fc7} features achieves 21.1 BLEU and 23.3\% ``Better or Equal to Human.'' Therefore, we believe that the 27.6\% achieved by the $k$-NN models is still relatively competitive with respect to the state-of-the-art.

However, given the strong BLEU/CIDEr performance of the nearest neighbor systems, this provides additional evidence that automatic metrics may only be a rough estimate of human judgments, as also noted in \cite{elliott2014comparing,hodosh2013framing,cider}.

\subsection{MS COCO Caption Test}

In Table \ref{tab:COCOtest}, we show results on {\tt fc7-fine} (CIDEr) on the MS COCO caption test set. Surprisingly, {\tt fc7-fine} is ranked second or third by most metrics. The METEOR metric computed using 40 captions per image (c40) ranks {\tt fc7-fine} fifth. As stated before, further human studies are still needed to gain a better understanding into how the captions produced by NN approaches are perceived by humans relative to generative approaches.

\section{Discussion}

The success of nearest neighbor approaches to image captioning
draws attention to the need for better evaluation and testing datasets. Ideally, we desire approaches that can generalize to images beyond those found in the training set. How can we build a testing set that measures this ability? One obvious approach is to measure the similarity of each testing image with those in the training set, similar to Figure \ref{fig:bins}. We could then examine how well approaches do on unusual or more diverse images. Another option would be to collect a new testing dataset using a different set of queries than those used for the MS COCO dataset. This would ensure the distribution of images in testing and training is different, and help us measure how well our approaches generalize.

The success of recent approaches such as \cite{ryan2014multimodal,mao2014explain,vinyals2014show,karpathy2014deep2,kiros2014unifying,donahue2014long,fang2014captions,chen2014learning,lebret2015phrase,lebret2014simple,lazaridou2015combining} demonstrates another problem with the task of image captioning. For each of these approaches, we know that human generated captions are typically preferred over the automatically generated captions. However, for many automatic evaluation metrics, the human captions have lower scores than the automatically generated captions. This suggests that the advancement towards  human-like captions may not be properly benchmarked using automatic approaches. Further research into automatic evaluation metrics that are highly correlated with human judgment is essential \cite{elliott2014comparing,hodosh2013framing,cider}.

A further difficulty when performing human evaluations on which caption is ``best'' is that we miss the nuances in the similarities/differences between the systems when we ask for humans' overall preferences. For example, we know that all of the NN captions are pretty fluent, while that may not be true for novel generated captions. However, we're not directly measuring fluency, so it is possible that generation-focused approaches are correctly capturing content, but are sometimes not fluent. On the other hand, NN approaches may produce very generic captions, causing them to not be preferred even when technically correct. Hopefully future experiments will shed light on these questions.

In this paper, we only explored very simple NN approaches to provide a baseline for the image captioning community. More sophisticated approaches that have been proposed for the task of caption ranking \cite{hodosh2013framing,mao2014explain,socher2013grounded,frome2013devise,karpathy2014deep,chen2014learning} may generate even better results. It may also be interesting to explore hybrid approaches that use NN approaches for query images with many similar training images and generation-based approaches for other images.


{\footnotesize
\bibliographystyle{ieee}
\bibliography{caption}
}

\end{document}